\pdfoutput=1
\documentclass[11pt]{article}
\usepackage{naacl2021}
\usepackage{times}
\usepackage{latexsym}
\usepackage[T1]{fontenc}
\usepackage[utf8]{inputenc}
\usepackage{microtype}
\usepackage{booktabs}

\usepackage{graphicx}
\usepackage{multirow}
\usepackage{adjustbox}
\usepackage{amsmath}
\usepackage{subcaption}
\usepackage{enumitem}
\usepackage[skip=5pt]{caption}
\usepackage{amsfonts}
\usepackage{mathrsfs}
\usepackage{amssymb}
\usepackage{amsthm}
\usepackage{fixmath}
\usepackage{tikz} 
\usepackage{float}
\usepackage{tabularx}
\usepackage{caption}
\usepackage{threeparttable}

\newcommand{\xhdr}[1]{{\noindent\bfseries #1} }
\newcommand{\methodname}{GraphMerge}

\definecolor{pengblue}{HTML}{4169E1}
\newcommand{\peng}[1]{{\color{pengblue}#1}}
\renewcommand{\peng}[1]{{#1}}

\title{Graph Ensemble Learning over Multiple Dependency Trees for Aspect-level Sentiment Classification}

\author{Xiaochen Hou\textsuperscript{$*$}, Peng Qi, Guangtao Wang, Rex Ying, Jing Huang,\\ \textbf {Xiaodong He, Bowen Zhou} \\
\\
\large{JD AI Research, Mountain View, California}\\
  {\tt \textsuperscript{$*$}xiaochen.hou1@jd.com} \\
}
\author{%
  Xiaochen Hou$^{1}$\textsuperscript{$*$}, Peng Qi$^{1}$, Guangtao Wang$^{1}$, Rex Ying$^{2}$, Jing Huang$^{1}$,\\ \textbf {Xiaodong He$^{1}$, Bowen Zhou$^{1}$}\\
$^{1}$JD AI Research, Mountain View, CA\\
$^{2}$Department of Computer Science, Stanford University, Stanford, CA\\
  {\tt \textsuperscript{$*$}xclmm1994@gmail.com}\\
}

\begin{document}
\maketitle
\begin{abstract}
Recent work on aspect-level sentiment classification has demonstrated the efficacy of incorporating syntactic structures such as dependency trees with graph neural networks (GNN), but these approaches are usually vulnerable to parsing errors.
To better leverage syntactic information in the face of unavoidable errors, we propose a simple yet effective graph ensemble technique, \methodname{}, to make use of the predictions from different parsers.
Instead of assigning one set of model parameters to each dependency tree, we first combine the dependency relations from different parses before applying GNNs over the resulting graph.
This allows GNN models to be robust to parse errors at no additional computational cost, and helps avoid overparameterization and
overfitting from GNN layer stacking by introducing more connectivity into the ensemble graph.
Our experiments on the SemEval 2014 Task 4 and ACL 14 Twitter datasets show that our \methodname{} model not only outperforms models with single dependency tree, but also beats other ensemble models without adding model parameters. 

\end{abstract}

\section{Introduction}
Aspect-level sentiment classification is a fine-grained sentiment analysis task, which aims to identify the sentiment polarity (e.g., positive, negative or neutral) of a specific aspect term in a sentence.
For example, in ``\textit{The exterior, unlike the food, is unwelcoming.}'', the polarities of aspect terms ``exterior'' and ``food'' are negative and positive, respectively.
This task has many applications, such as assisting customers to filter online reviews or make purchase decisions on e-commerce websites.

Recent studies have shown that syntactic information such as dependency trees is very effective in capturing long-range syntactic relations that are obscure from the surface form~\cite{zhang2018graph}. 
Several successful approaches employed graph neural network (GNN)~\cite{kipf2016semi} model over dependency trees to  aspect-level sentiment classification~\cite{huang2019syntax,zhang2019aspect,sun2019aspect,wang2020relational}, which demonstrate that syntactic information is helpful for associating the aspect term with relevant opinion words more directly for increased robustness in sentiment classification. 

\begin{figure}
\centering
\includegraphics[width=\linewidth]{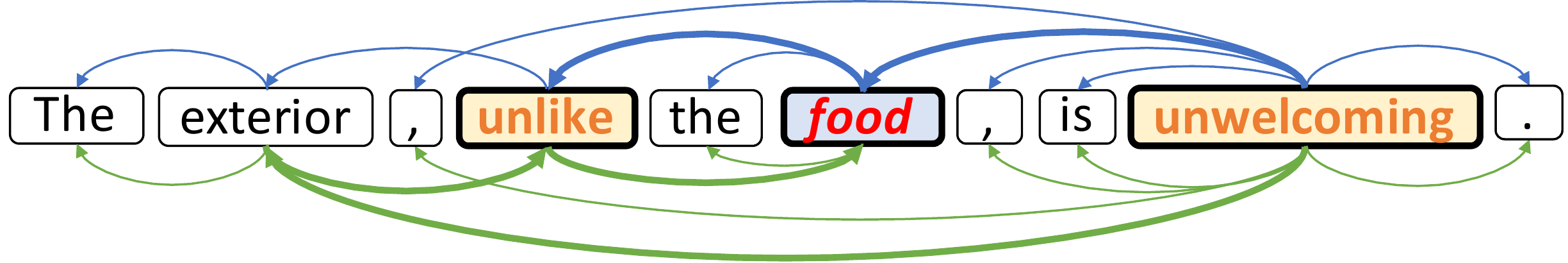}
\caption{
An example where an incorrect parse (above the sentence) can mislead aspect-level sentiment classification for the term ``food'' by connecting it to the negative sentiment word ``unwelcoming'' by mistake.
Although having its own issues, the parse below correctly captures the main syntactic structure between the aspect terms ``exterior'', ``food'' and the sentiment word, and is more likely to lead to a correct prediction.
}
\label{fig:wrong}
\end{figure}

However, existing approaches are vulnerable to parsing errors~\cite{wang2020relational}.
For example, in Figure~\ref{fig:wrong}, the blue parse above the sentence can mislead models to predict negative sentiment for the aspect term ``food'' with its direct association to ``unwelcoming''.
Despite their high edge-wise parsing performance on standard benchmarks, state-of-the-art dependency parsers usually struggle to predict flawless parse trees especially in out-of-domain settings.
This poses great challenge to dependency-based methods that rely on these parse trees---the added benefit from syntactic structure does not always prevail the noise introduced by model-predicted parses ~\cite{he2017deep, sachan2020syntax}.


In this paper, we propose \methodname{}, a \emph{graph ensemble} technique to help dependency-based models mitigate the effect of parsing errors.
Our technique is based on the observation that different parsers, especially ones with different inductive biases, often err in different ways.
For instance, in Figure 1, the green parse under the sentence is incorrect around ``unlike the food'', but it nevertheless correctly associates ``unwelcoming'' with the other aspect term ``exterior'', and therefore is less likely to mislead model predictions.
Given dependency trees from multiple parses, instead of assigning each dependency tree a separate set of model parameters and ensembling model predictions or dependency-based representations of the same input, we propose to combine the different dependency trees before applying representation learners such as GNNs.

Specifically, we take the union of the edges in all dependency trees from different parsers to construct an ensemble graph, before applying GNNs over it.
This exposes the GNN model to various graph hypotheses at once, and allows the model to learn to favor edges that contribute more to the task.
To retain the syntactic dependency information between words in the original dependency trees, we also define two different edge types---parent-to-children and children-to-parent---which are encoded by applying relational graph attention networks (RGAT)~\cite{busbridge2019relational} on the ensemble graph.

Our approach has several advantages. Firstly, since \methodname{} combines dependency trees from different parsers, the GNN models can be exposed to multiple parsing hypotheses and learn to choose edges that are more suitable for the task from data.
As a result, the model is less reliant on any specific parser and more robust to parsing errors.
Secondly, this improved robustness to parsing errors at no additional computational cost, since we are still applying GNNs to a single graph with the same number of nodes.
Last but not least, \methodname{} helps prevent GNNs from overfitting by limiting over-parameterization.
Aside from keeping the GNN computation over a single graph to avoid separate parameterization for each parse tree, \methodname{} also introduces more edges in the graph when parses differ, which reduces the diameter of graphs.
As a result, fewer layers of GNNs are needed to learn good representations from the graph, alleviating the over-smoothing problem~\cite{li2018deeper}.


To summarize, the main contribution of our work are the following:
\begin{itemize}[leftmargin=6pt]
\setlength{\itemsep}{0pt}%
    \setlength{\parskip}{2pt}
    \item We propose a \methodname{} technique to combine dependency parsing trees from different parsers to improve model robustness to parsing errors. The ensemble graph enables the model to learn from noisy graph and select correct edges among nodes at no additional computational cost.
    \item 
    We retain the syntactic dependency information in the original trees by parameterizing parent-to-children and children-to-parent edges separately, which improves the performance of the RGAT model on the ensemble graph.
    \item 
    Our \methodname{} RGAT model outperforms recent state-of-the-art work on three benchmark datasets (Laptop and Restaurant reviews from SemEval 2014 and the ACL 14 Twitter dataset). It also outperforms its single-parse counterparts as well as other ensemble techniques.
\end{itemize}

\begin{figure*}
\centering
\includegraphics[width=.8\linewidth]{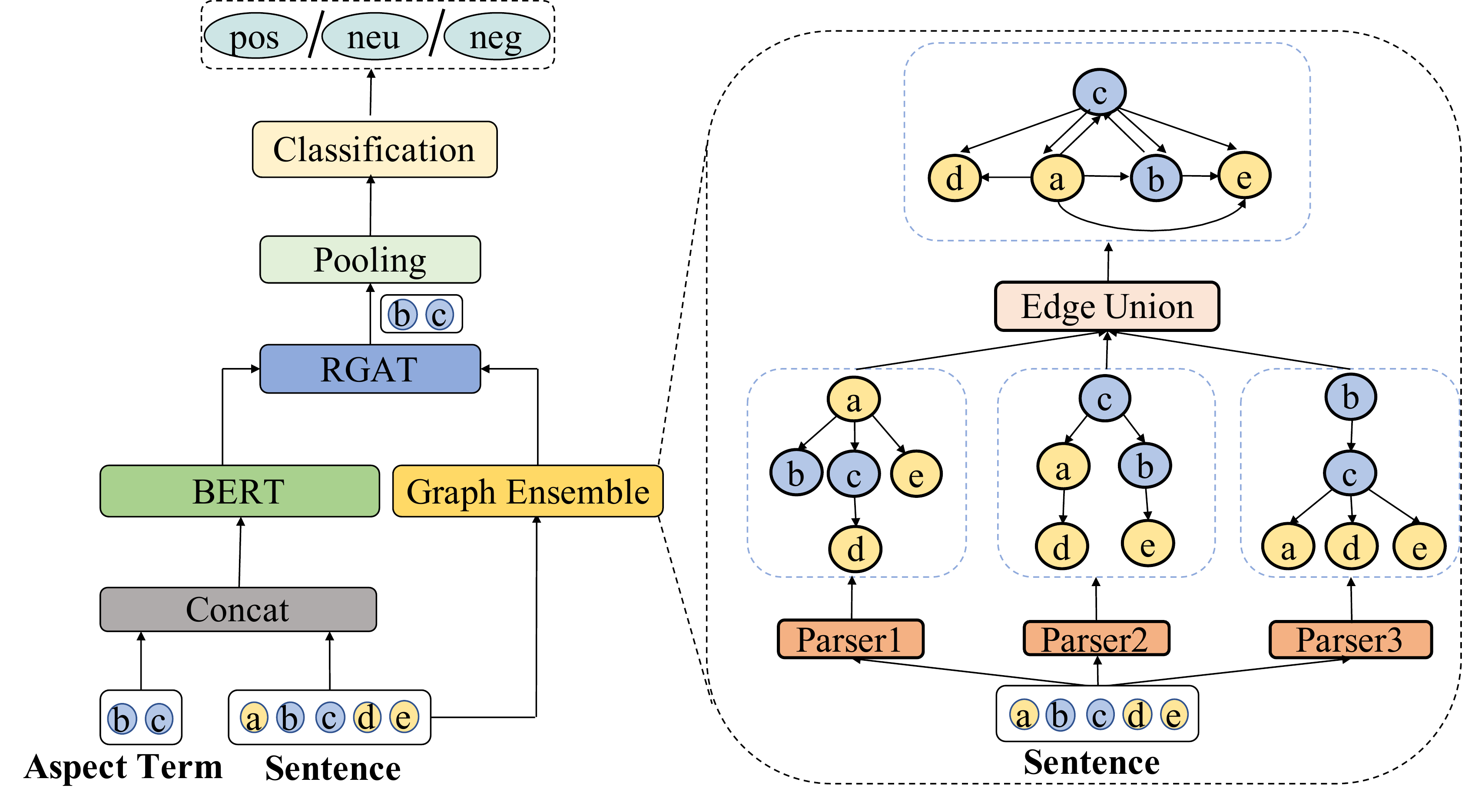}
\caption{The framework of the \methodname{} model for aspect-level sentiment classification over multiple dependency trees. The left side shows the overall architecture for sentiment classification; and the right side shows the details of how to perform graph ensemble with \methodname{}.}
\label{fig:model}
\end{figure*}

\section{Related Work}
Much recent work on aspect-level sentiment classification has focused on applying attention mechanisms (e.g., co-attention, self attention, and hierarchical attention) to sequence models such recurrent neural networks (RNNs)
~\cite{tang2015effective,tang2016aspect,liu2017attention,wang2018target,fan2018multi,chen2017recurrent,zheng2018left,wang2018learning,li2018hierarchical,li2018transformation}. 
In a similar vein, pretrained transformer language models such as BERT~\cite{devlin2018bert} have also been applied to this task, which operates directly on word sequences~\cite{song2019attentional,xu2019bert,rietzler2019adapt}.

In parallel, researchers have also found syntactic information to be helpful for this task, and incorporated it into aspect-level sentiment classification models in the form of dependency trees~\cite{dong2014adaptive, he2018effective} as well as constituency trees~\cite{nguyen2015phrasernn}.
More recently, researchers have developed robust dependency-based models with the help of GNNs that operate either directly on dependency trees~\cite{huang2019syntax, zhang2019aspect, sun2019aspect}, as well as reshaped dependency trees that center around aspect terms~\cite{wang2020relational}.
While most recent work stack GNNs on top of BERT models, \citet{tang-etal-2020-dependency} have also reported gains by jointly learning the two with a mutual biaffine attention mechanism.

Despite the success of these dependency-based models, they are usually vulnerable to parse errors since they rely on a single parser.
Furthermore, most prior work that leverage GNNs to encode dependency information treats the dependency tree as an undirected graph, therefore ignores the syntactic relation between words in the sentence.


\section{Proposed Model}
We are interested in the problem of predicting the sentiment polarity of an aspect term in a given sentence.
Specifically, given a sentence of $n$ words $\{w_1,w_2,\ldots,w_{\tau}, \ldots,w_{\tau+t}, \ldots,w_n\}$ where $\{w_{\tau}, w_{\tau+1},\ldots,w_{\tau+t-1}\}$ is the aspect term, the goal is to classify the sentiment polarity toward the term as positive, negative, or neutral. 
Applying GNNs over dependency trees is shown effective to solve this problem, however it is vulnerable to parsing errors. Therefore, we propose a \methodname{} technique to utilize multiple dependency trees to improve robustness to parsing errors.
In this section, we will first introduce \methodname{}, our proposed graph ensemble technique, then introduce the GNN model over \methodname{} graph for aspect-level sentiment analysis.

\subsection{\methodname{} over Multiple Dependency Trees}

To allow graph neural networks to learn dependency-based representations of words while being robust to parse errors that might occur, we introduce \methodname{}, which combines different parses into a single ensemble graph.
Specifically, given a sentence $\{w_1,w_2,\ldots,w_n\}$ and $M$ different dependency parses $G_1, \ldots, G_M$, \methodname{} takes the union of the edges from all parses, and constructs a single graph $G$ as follows
\begin{flalign}
 G &= (V, \{e | e = (w_i, w_j) \in \bigcup_{m=1}^M E_m\})
\end{flalign}
where $V$ is the shared set of nodes among all graphs\footnote{This is true for dependency trees as long as parsers share the same tokenization as input.} and $E_m (1 \leq m \leq M)$ is the set of edges in $G_m$ (see the right side of Figure \ref{fig:model} for an example).



As a result, $G$ contains all of the (directed) edges from all dependency trees, on top of which we can apply the same GNN models when a single dependency tree is used.
Therefore, \methodname{} introduces virtually no computational overhead to existing GNN approaches, compared to traditional ensemble approaches where computational time and/or parameter count scale linearly in $M$.
What is more, the resulting graph $G$ likely contains more edges from the gold parse which correctly captures the syntactic relation between words in the sentence, allowing the GNN to be robust to parse errors from any specific parser.
Finally, since $G$ contains more edges between words when parses differ than any single parse and reduces the diameter of the graph, it is also more likely that a shallower GNN model is enough 
to learn good representations, therefore avoiding over-parameterization and thus overfitting from stacking more GNN layers.

\subsection{RGAT over Ensemble Graph}

\peng{To learn node representations from ensemble graphs, we apply graph attention networks~\cite[GAT;][]{velivckovic2017graph}}.
In one layer of GAT, the hidden representation of each node in the graph is computed by attending over its neighbors, with a multi-head self-attention mechanism. 
The representation for word $i$ at the $l$-th layer of GAT can be obtained as follows
\begin{flalign}
\mathbold{h}^{(l)}_i&=\mathbin\Vert_{k=1}^{K}\sigma(\sum_{j \in N_i}\alpha_{ij}^k\mathbold{W}^k\mathbold{h}^{(l-1,k)}_i)
\end{flalign}
Where $K$ is the number of attention heads, $N_i$ is the neighborhood of node $i$ in the graph, and $\mathbin\Vert$ the concatenation operation. $\mathbold{W}^k \in \mathbb{R}^{d_B \times d_h}$ represents the learnable weights in GAT and $\sigma$ denotes $\mathrm{ReLU}$ activation function. $\alpha_{ij}^k$ is the attention score between node $i$ and node $j$ with head $k$. 

\xhdr{Edge Types.}
\peng{To apply GAT to ensemble graphs, we first add reciprocal edges for each edge in the dependency tree, and label them with parent-to-children and children-to-parent types, respectively.
This allows our model to retain the original syntactic relation between words in the sentence.
We also follow previous work to add self loop to each node in the graph, which we differentiate from dependency edges by introducing a third edge type.}

\peng{We adapt Relational GAT (RGAT) to capture this edge type information.}
Specifically, we encode the edge type information when computing the attention score between two nodes. 
\peng{We assign each edge type an embedding $\mathbold{e} \in \mathbb{R}^{d_h}$, incorporate it into attention score computation as follows}
\begin{flalign}
\alpha_{ij}&=\frac{\exp(\sigma(\mathbold{a}\mathbold{W}(\mathbold{h}_i\|\mathbold{h}_j)+\mathbold{a}_e\mathbold{e}_{ij}))}{\sum_{v \in N_i}\exp(\sigma(\mathbold{a}\mathbold{W}(\mathbold{h}_i\|\mathbold{h}_v)+\mathbold{a}_e\mathbold{e}_{iv}))} \label{eq2}
\end{flalign}
where $\mathbold{e}_{ij}$ is the representation of the type of the edge connecting nodes $i$ and $j$. $\mathbold{a} \in \mathbb{R}^{d_h}$, $\mathbold{W} \in \mathbb{R}^{d_h \times 2d_h}$ and $\mathbold{a}_e \in \mathbb{R}^{d_h}$ are learnable matrices. 


\subsection{Sentiment Classification}
We extract hidden representations from nodes that correspond to aspect terms in the last RGAT layer, and conduct average pooling to obtain $\mathbold{h}_t \in \mathbb{R}^{d_{h}}$. Then we feed it into a two-layer MLP to calculate the final classification scores $\hat{\mathbold{y}}_s$:
\begin{flalign}
\hat{\mathbold{y}}_s&=\mathrm{softmax}({\mathbold{W}_{2}\mathrm{ReLU}({\mathbold{W}_{1}\mathbold{h}_{t}})})
\end{flalign}
where $\mathbold{W}_{2} \in \mathbb{R}^{C \times d_{out}}$ and $\mathbold{W}_{1} \in \mathbb{R}^{d_{out} \times d_{h}}$ denote  learnable weight matrices, and $C$ is the number of sentiment classes.
\peng{We optimize the model to minimize the standard cross entropy loss function, and apply weight decay to model parameters.}

\subsection{RGAT Input}

The initial word node features for RGAT are obtained from a BERT encoder, with positional information from positional embeddings. 

\xhdr{BERT Encoder.}
We use the pre-trained BERT base model as the encoder to obtain word representations. Specifically, we construct the input as ``[CLS] + sentence + [SEP] + term + [SEP]'' and feed it into BERT. 
\peng{This allows BERT to learn term-centric representations from the sentence during fine-tuning.
To feed the resulting wordpiece-based representations into the word-based RGAT model, we average pool representations of subwords for each word to obtain $\mathbold{X}$, the raw input to RGAT.}

\xhdr{Positional Encoding.}
Position information is beneficial 
for this task, especially when there are multiple aspect terms in one sentence, where it  helps to locate opinion words relevant to an aspect term. Although the BERT encoder already takes the word position into consideration, 
\peng{it is dampened after layers of Transformers.}
Therefore, we \peng{explicitly} encode the absolute position for each word \peng{and add it to the BERT output}. 
\peng{Specifically, we add a trainable position embedding matrix to $\mathbold{X}$ before feeding the resulting representation into RGAT.}

\begin{table}
\centering
\small
\begin{adjustbox}{max width=0.5\textwidth}
\begin{tabular}{ c c c c c c c }
\toprule
\multirow{2} {*}{Dataset} & \multicolumn{2}{c}{Positive} & \multicolumn{2}{c}{Neutral} & \multicolumn{2}{c}{Negative}\\
\cline{2-7}
& Train & Test & Train & Test & Train & Test\\
\hline
Laptop &  \phantom{0}987 & 341 & \phantom{0}460 & 169 & \phantom{0}866 & 128\\
Restaurant & 2164 & 728 & \phantom{0}633 & 196 & \phantom{0}805 & 196\\
Twitter & 1561 & 173 & 3127 & 346 & 1560 & 173 \\
\bottomrule
\end{tabular}
\end{adjustbox}
\caption{Statistics of \peng{the three benchmark datasets used in our experiments}.}
\label{table:data}
\end{table}

\begin{table*}[!ht]
\centering
\resizebox{1.9\columnwidth}{!}{
\begin{threeparttable}
\begin{tabularx}{1.15\textwidth}{c c c c c c c c}
\toprule
\multirow{2} {*}{Category}  &\multirow{2} {*}{Model}  & \multicolumn{2}{c}{14Rest} & \multicolumn{2}{c}{14Lap} & \multicolumn{2}{c}{Twitter} \\
\cline{3-8}
 & & Acc & Macro-F1 & Acc & Macro-F1 & Acc & Macro-F1 \\
\midrule
\multirow{2} {*}{BERT}&BERT-SPC~\cite{song2019attentional} 
& 84.46 & 76.98 & 78.99 & 75.03 & 73.55 & 72.14 \\
&AEN-BERT~\cite{song2019attentional} 
& 83.12 & 73.76 & 79.93 & 76.31 & 74.71 & 73.13  \\
\hline
BERT+DT\tnote{$\star$} &DGEDT-BERT~\cite{tang-etal-2020-dependency}  & 86.3 & 80.0 & 79.8 & 75.6 & 77.9 & 75.4\\
\hline
BERT+RDT{$^\diamond$} & R-GAT+BERT~\cite{wang2020relational} & 86.60  & 81.35  & 78.21 & 74.07 & 76.15 & 74.88 \\
\hline
Ours &{\textbf{\methodname{}}} & \textbf{87.32} & \textbf{81.95} &  \textbf{81.35} &  \textbf{78.65} & \textbf{78.18} & \textbf{76.52}\\
\bottomrule
\end{tabularx}
		\begin{tablenotes}
			 \item $\star$ DT: Dependency Tree; $\diamond$ RDT: Reshaped Dependency Tree.
		\end{tablenotes}
\end{threeparttable}
}
\caption{Comparison of our \methodname{} model to different published numbers on three datasets, with the same setup of train and test data -- no dev data. The bold text indicates the best results.}
\label{table:bestresults}
\end{table*}

\section{Experiments}
\xhdr{Data \& Processing.} We evaluate our model on three datasets: Restaurant and Laptop reviews from SemEval 2014 Task 4 \peng{(14Rest and 14Lap)}\footnote{https://alt.qcri.org/semeval2014/task4/} and ACL 14 Twitter dataset \peng{(Twitter)}~\cite{dong2014adaptive}.
We remove several examples with ``conflict'' sentiment polarity labels in the reviews.  The statistics of these datasets are listed in Table \ref{table:data}. 
\peng{Following previous work, we report the accuracy and macro F1 scores for sentiment classification.

For dependency-based approaches, we tokenize sentences with Stanford CoreNLP~\cite{manning2014stanford}, and then parse them with CoreNLP, Stanza~\cite{qi2020stanza}, and the Berkeley neural parser~\cite{Kitaev-2018-SelfAttentive}.
Since the Berkeley parser returns constituency parses, we further convert it into dependency parses using CoreNLP.}

\xhdr{Baselines.} 
We compare our \methodname{} model against published work on these benchmarks, including:
\textbf{BERT-SPC}~\cite{song2019attentional} feeds the sentence and term pair into the BERT model and uses the BERT outputs for predictions; \textbf{AEN-BERT}~\cite{song2019attentional} uses BERT as the encoder and employs several attention layers. 
BERT + Dependency tree based models: \textbf{DGEDT-BERT}~\cite{tang-etal-2020-dependency} proposes a mutual biaffine module to jointly consider the representations learnt from Transformer and the GNN model over the dependency tree;
\textbf{R-GAT+BERT}~\cite{wang2020relational} reshapes and prunes the dependency tree to an aspect-oriented tree rooted at the aspect term, and then employs RGAT to encode the new tree for predictions.
For fair comparison, we report the results of our \methodname{} model using the same data split (without a development set).

To understand the behavior of different models, we also implement several baseline models. In our experiments, we randomly sample 5\% training data as held-out development set for hyper-parameter tuning, use the remaining 95\% for training and present results of the average and standard deviation numbers from five runs of random initialization on the test set. We consider these baselines:
\begin{enumerate}[itemsep=1pt,leftmargin=10pt]
    \item \textit{BERT-baseline} which feeds the sentence-term pair into the BERT-base encoder and then applies a  classifier \peng{with the representation of the aspect term token}.
    \item \textit{GAT-baseline with Stanza} which employs a vanilla GAT model over single dependency tree obtained from Stanza without differentiating edge types. And the initial node features are the raw output of the BERT encoder.
    \item 
    \textit{RGAT over single dependency trees}, where we apply RGAT models with parent-to-children and child-to-parent edge types over different dependency trees from the CoreNLP, Stanza, and Berkeley parsers. For a fair comparison to our \methodname{} model, the RGAT input comes from BERT encoder plus position embeddings.
    \item Two ensemble models to take advantage of multiple dependency trees, including a \textit{Label-Ensemble} model which takes the majority vote from three models each trained on one kind of parses, and a \textit{Feature-Ensemble} model which applies three sets of RGAT parameters, one for each parse, on top of the BERT encoder with their output features concatenated.
    These models have more parameters and are more computationally expensive compared to the \methodname{} model when operating on the same parses.
\end{enumerate}




\xhdr{Parameter Setting.} We use Pytorch~\cite{NEURIPS2019_9015} to implement our models. The GAT implementation is based on Deep Graph Library~\cite{wang2019dgl}. During training, we set the learning rate = $10^{-5}$, batch size = $4$. 
We use dev data to select the hidden dimension $d_{h}$ for GAT/RGAT from $\{64, 128, 256\}$, the head number in the multi-head self-attention from $\{4,8\}$, and GAT/RGAT layer from $\{2,3,4\}$. The 2-layer GAT/RGAT models turn out to be the best based on the dev set. We apply dropout~\cite{srivastava2014dropout} and select the best setting from the dropout rate range = $[0.1,0.3]$. We set the weight of L2 regularization as $10^{-6}$. We train the model up to 5 epochs.\footnote{Our code will be released at the time of publication.}

\subsection{Experimental Results}

\begin{table*}
\centering
\resizebox{2.0\columnwidth}{!}{
\begin{tabular}{c c c c c c c}
\toprule
\multirow{2} {*}{Model}  & \multicolumn{2}{c}{14Rest} & \multicolumn{2}{c}{14Lap} & \multicolumn{2}{c}{Twitter} \\
\cline{2-7}
  & Acc & Macro-F1 & Acc & Macro-F1 & Acc & Macro-F1 \\
\midrule
 BERT-baseline & 83.43 $\pm$ 0.52 & 74.94 $\pm$ 1.37 & 77.34 $\pm$ 0.90 & 72.77 $\pm$ 1.96 & 73.47 $\pm$ 0.89 & 72.63 $\pm$ 0.82\\
{GAT-baseline with Stanza} & 84.29 $\pm$ 0.30 & 75.75 $\pm$ 1.11 & 77.84 $\pm$ 0.27 & 74.0 $\pm$ 0.55 & 73.82 $\pm$ 0.70 & 72.72 $\pm$ 0.42 \\
\midrule
{RGAT with Stanza} & 84.53 $\pm$ 0.66 & 77.29 $\pm$ 1.42 & 77.99 $\pm$ 0.62 & 74.35 $\pm$ 0.60 & 73.99 $\pm$ 0.48 & 72.76 $\pm$ 0.33 \\
{RGAT with Berkeley} & 84.41 $\pm$ 0.86 & 76.63 $\pm$ 1.38 & 78.09 $\pm$ 1.24 & 73.65 $\pm$ 1.76 & 74.07 $\pm$ 0.67 & 72.65 $\pm$ 0.74 \\
{RGAT with CoreNLP} & 83.86 $\pm$ 0.32 & 76.03 $\pm$ 0.88 & 78.12 $\pm$ 1.02 & 73.86 $\pm$ 1.72 & 73.96 $\pm$ 0.93 & 72.83 $\pm$ 0.95 \\
\midrule
{Label-Ensemble} & 84.68 $\pm$ 0.95 & 77.21 $\pm$ 1.54 & 78.40 $\pm$ 1.51 & 74.36 $\pm$2.45 & 74.59 $\pm$ 0.46 & 73.51 $\pm$ 0.43 \\
{Feature-Ensemble} & 84.64 $\pm$ 0.77 & 77.06 $\pm$ 1.45 & 78.68 $\pm$ 0.69 & 74.80 $\pm$ 0.92 & 74.62 $\pm$ 0.76 & 73.61 $\pm$ 0.73 \\
\midrule
{\textbf{\methodname{}}} & \textbf{85.16} $\pm$ 0.53 &  \textbf{77.91} $\pm$ 0.87 &  \textbf{80.00} $\pm$ 0.63 &  \textbf{76.50} $\pm$ 0.64 &  \textbf{74.74} $\pm$ 0.93 &  \textbf{73.66} $\pm$ 0.88 \\
\bottomrule
\end{tabular}
}
\caption{Comparison of our \methodname{} model to different baselines on three datasets, with 5\% dev data set aside. The bold text indicates the best results.}
\label{table:results}
\end{table*}

\begin{table*}[!h]
\centering
\resizebox{2\columnwidth}{!}{
\begin{threeparttable}
\begin{tabularx}{1.17\textwidth}{l c c c c c c}
\toprule
\multirow{2} {*}{Model}  & \multicolumn{2}{c}{14Rest} & \multicolumn{2}{c}{14Lap} & \multicolumn{2}{c}{Twitter}\\
\cline{2-7}
& Acc & Macro-F1 & Acc & Macro-F1 & Acc & Macro-F1\\
\midrule
\textbf{\methodname{}} & \textbf{85.16} $\pm$ 0.53 &  \textbf{77.91} $\pm$ 0.87 &  \textbf{80.00} $\pm$ 0.63 &  \textbf{76.50} $\pm$ 0.64 &  \textbf{74.74} $\pm$ 0.93 &  \textbf{73.66} $\pm$ 0.88 \\
\midrule
- Edge type & 84.25 $\pm$ 0.59 & 76.15 $\pm$ 1.24 & 78.65 $\pm$ 0.51 & 74.76 $\pm$ 0.71 & 74.37 $\pm$ 1.08 & 73.25 $\pm$ 0.85 \\
- Position & 84.36 $\pm$ 0.36 & 75.92 $\pm$ 1.18 & 78.37 $\pm$ 0.31 & 74.51 $\pm$ 0.48 & 74.28 $\pm$ 1.39 & 73.34 $\pm$ 1.35\\
- (Edge type + Position) & 84.16 $\pm$ 0.31 & 75.38 $\pm$ 0.69 & 78.09 $\pm$ 0.27 & 74.29 $\pm$ 0.64 & 73.41 $\pm$ 0.63 & 72.52 $\pm$ 0.62\\
\midrule
\peng{Edge Intersection} & 84.59 $\pm$ 0.61 & 77.06 $\pm$ 1.07 & 78.65 $\pm$ 0.94 & 74.86 $\pm$ 1.42 & 74.68 $\pm$ 0.83 & 73.45 $\pm$ 0.73\\
\bottomrule
\end{tabularx}
\end{threeparttable}
}
\caption{Ablation study of the \methodname{} model over three datasets. We report the average and standard deviation over five runs, where Edge Intersection means taking intersection of edges from multiple dependency trees.}
\label{table:ablation}
\end{table*}
We first compare our model to previous work following the evaluation protocol in previous work, and report results in Table \ref{table:bestresults}. As we can see, the \methodname{} model achieves best performances on all three datasets. 
On the Laptop dataset, the \methodname{} model further outperforms baselines by at least 1.42 accuracy and 2.34 Macro-F1 respectively. 

Table \ref{table:results} shows performance comparisons of the \methodname{} model with other baselines in terms of accuracy and Macro-F1. We observe that:

\xhdr{Syntax information benefits aspect-level sentiment classification.} 
All GAT and RGAT models based on dependency trees outperform \textit{BERT-baseline} on all three datasets. This demonstrates that leveraging syntax structure information is beneficial to this task. 

\xhdr{Ensemble models benefit from multiple parses.} The \textit{Label-Ensemble}, \textit{Feature-Ensemble}, and \methodname{} models achieve better performance compared to their single dependency tree counterparts. 
This shows that ensemble models benefit from the presence of different parses and thus less sensitive to parse errors from any single parser.

\xhdr{\methodname{} achieves the best performance overall.}Our proposed \textit{\methodname{}} model not only shows consistent improvements over all single dependency tree models, but also surpasses the other two ensemble models without additional parameters or computational overhead, when compared to the single-tree models.



\subsection{Model Analysis}
We analyze the proposed \methodname{} model from two perspectives: an ablative analysis of model components and an analysis of the change in the dependency graphs after \methodname{} is applied.

\subsubsection{Ablation Study}
\xhdr{Model components.}
We conduct ablation studies of our modeling for edge type and position information in Table \ref{table:ablation}. We observe that: (1) On three datasets, ablating the edge type degrades the performances. It indicates that the syntactic dependency information in original dependency trees is important. Differentiating edges in the ensemble graph provides more guidance to the model about selecting useful connections among nodes. (2) Removing the position embeddings hurts the performances as well. Although the BERT encoder already incorporates  position information at its input, this information is dampened over the layers of Transformers. 
Emphasizing sequence order again before applying RGAT benefits the task. 

\xhdr{Edge Union vs. Edge Intersection.}
While \methodname{} keeps all edges from different dependency parsing trees for the RGAT model to learn to use, this could also result in too much structural noise and adversely impact performance.
We therefore compare \methodname{} to edge intersection, which only retains edges that shared by all individual trees when constructing the ensemble graph, which can be thought of distilling syntactic information that an ensemble parser is confident about. 
We observe from the last row in Table \ref{table:ablation} that edge intersection strategy underperforms \methodname{} on all datasets. 
We postulate that this is because edge intersection over-prunes edges in the ensemble graph and might introduce more disjoint connected components where parsers disagree, which the RGAT model cannot easily recover from.



\begin{table}
\centering
\small
\begin{adjustbox}{max width=0.475\textwidth}
\begin{tabular}{ c c c c }
\toprule
& Aspect Terms & Opinion Words & Coverage\\
\midrule
Laptop & \phantom{0}638 & 467 & 73.20\%\\
Restaurant & 1120 & 852 & 76.07\%\\
\bottomrule
\end{tabular}
\end{adjustbox}
\caption{Statistics of the Opinion datasets. Aspect Terms denotes as the total number of aspect terms. Opinion Words represents total number of terms that have labeled opinion words. Coverage is the proportion of terms with labeled opinion words.}
\label{table:opinion_data}
\end{table}
\begin{figure}[t]
\centering
\begin{subfigure}[h]{.5\textwidth}
\centering
\includegraphics[width=0.95\linewidth]{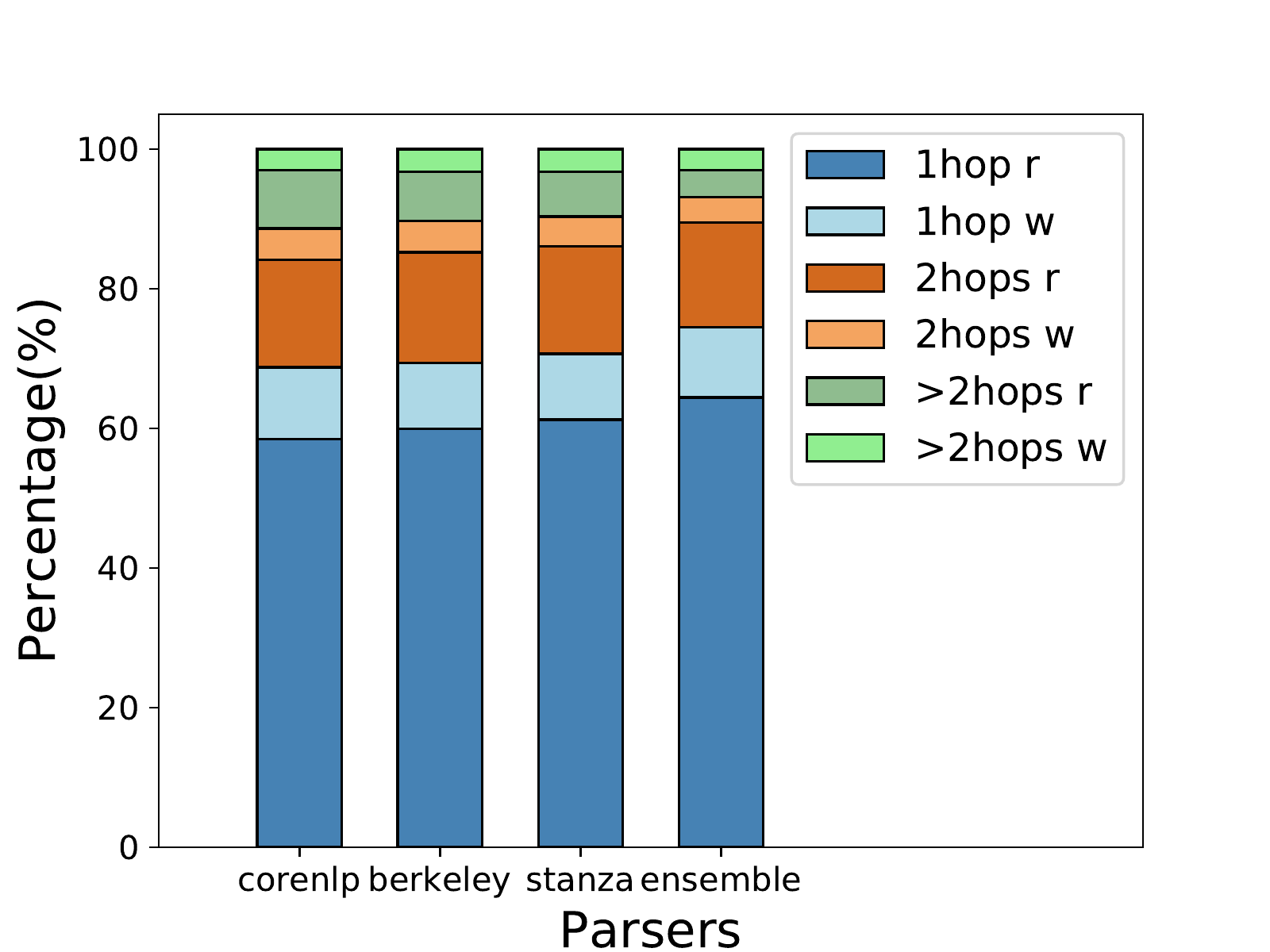}
\caption{Accuracy w.r.t different hop number on 14Lap.}
\end{subfigure}
\begin{subfigure}[h]{.5\textwidth}
\centering
\includegraphics[width=0.95\linewidth]{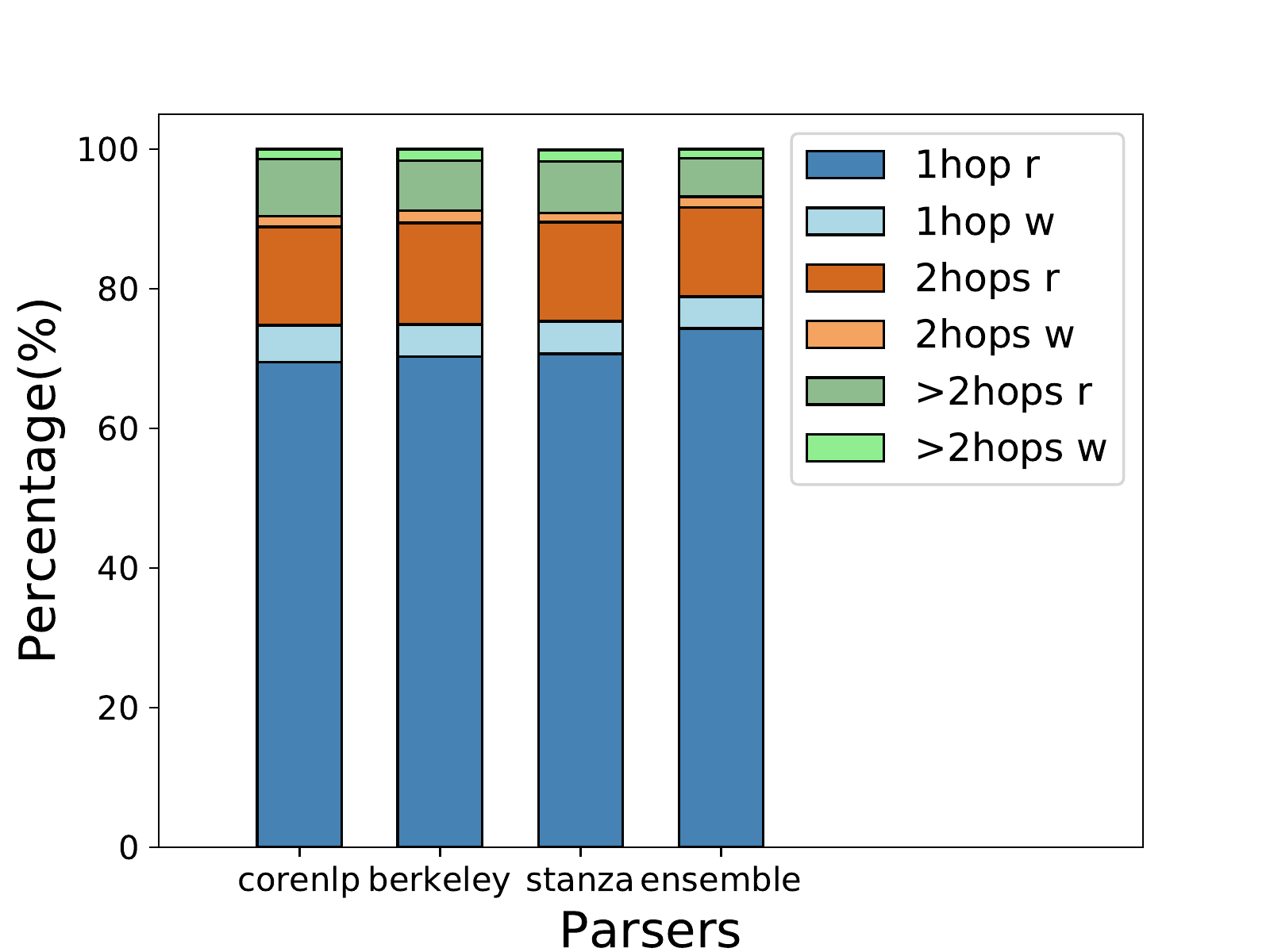}  
\caption{Accuracy w.r.t different hop number on 14Rest.}
\end{subfigure}
\caption{Hop analysis on 14Lap and 14Rest, where ``r'' and ``w'' denote the right and wrong prediction, respectively.}
\label{fig:hopanalysis}
\end{figure}

\subsubsection{Graph Structure Analysis}
\xhdr{Effect of GraphMerge on Graph Structure.} 
To better understand the effect of \methodname{} on dependency graphs, we conduct statistical analysis on the test set of 14Lap and 14Rest.
Specifically, we are interested in the change in the shortest distance between the aspect term and its opinion words on the dependency graphs.
For this analysis, we use the test sets with opinion words labeled by \citet{fan-etal-2019-target} (see Table \ref{table:opinion_data} for dataset statistics).

We summarize analysis results in Figure \ref{fig:hopanalysis}. We observe that: (1) Compared with single dependency tree, the ensemble graph effectively increases the number of one-hop and two-hops cases, meaning the overall distance between the term and opinion words is shortened on both datasets. (2) Shorter distance between the term and opinion words correlates with better performance. With the ensemble graph, the accuracy of one-hop and two-hops cases beats all single dependency tree models. 
These observations suggest that the ensemble graph from \methodname{} introduces important connectivity to help alleviate overparameterization from stacking RGAT layers, and that the RGAT model is able to make use of the diversity of edges in the resulting graph to improve classification performance.

Note that although shortening distance correlates with improved results, it does not mean that the closer distance is sufficient for better performance. 
This is because although the BERT model can be seen as a GAT over a fully-connected graph where a word is reachable for all other context words within one hop~\cite{wang2020direct}, the \textit{BERT-baseline} model performs worse than dependency-based models. Therefore, encoding the syntactic structure information in dependency trees is crucial for this task. Our \methodname{} model achieves the best results by shortening the graph distance between the aspect term and opinion words with syntactic information.

\begin{figure*}[!h]
\centering
\includegraphics[width=0.98\linewidth]{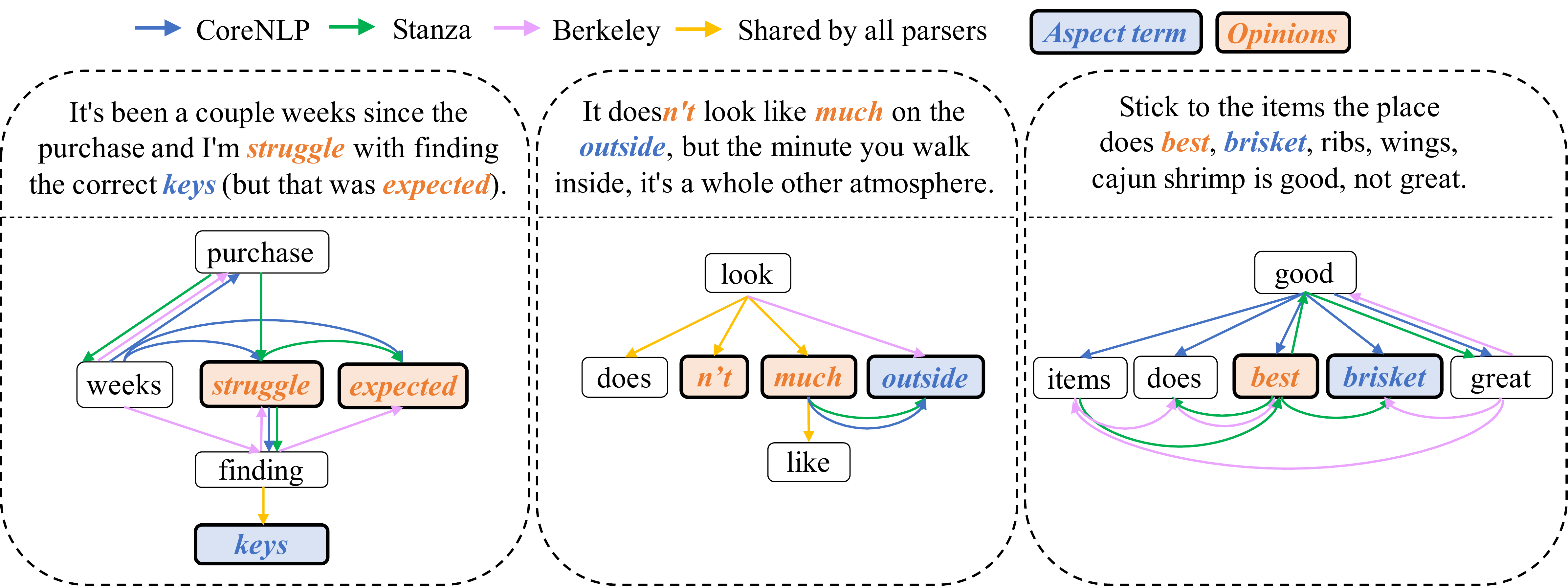}
\caption{Examples of partial dependency trees on which the single dependency tree models make wrong prediction, but the \methodname{} model makes correct prediction.}
\label{fig:casestudy}
\end{figure*}
\xhdr{Case Study.} 
To gain more insight into the \methodname{} model's behaviour, we find several examples and visualize their dependency trees from three parsers (Figure \ref{fig:casestudy}). Due to the space limit, we only show partial dependency trees that contain essential aspect terms and opinion words. These examples are selected from cases that all single dependency tree RGAT models predict incorrectly, but the \textit{\methodname{}} model predicts correctly. 


We observe that in general, the three parsers do not agree in the neighborhood around the aspect term and opinion words in these sentences.
As a result, \methodname{} tends to shorten the distance between the aspect term and the opinion words on the resulting graph.
For instance, for all examples in Figure \ref{fig:casestudy}, the shortest distances between the aspect term and the opinion words are no more than two in the ensemble graphs, while they vary from 2 to 4 in the original parse trees.
This could allow the RGAT model to capture the relation between the words without an exessive amount of layers, thus avoiding overfitting.

On the other hand, we observe that the resulting ensemble graph from \methodname{} is more likely to contain the gold parse for the words in question. For instance, in the first two examples, the gold parse for the words visualized in the figure can be found in the ensemble graph (despite no individual parser predicting it in the first example); the third example also has a higher recall of gold parse edges than each parser despite being difficult to parse.
This offers the RGAT model with the correct semantic relationship between these words in more examples during training and evaluation, which is often not accessible with those single parse trees.


\begin{table}
\centering
\small
\begin{adjustbox}{max width=0.475\textwidth}
\begin{tabular}{ c c c c c }
\toprule
Dataset & Positive & Neutral & Negative & Total \\
\midrule
Laptop & \phantom{0}883 & 407 & 587 & 1877 \\
Restaurant & 1953 & 473 & 1104 & 3530\\
\bottomrule
\end{tabular}
\end{adjustbox}
\caption{Statistics of robustness testing data ARTS.}
\label{table:RobustnessData}
\end{table}
\begin{table}
\centering
\resizebox{1.0\columnwidth}{!}{
\begin{threeparttable}
\begin{tabularx}{0.75\textwidth}{l c c }
\toprule
\multirow{2} {*}{Model}  & 14Rest & 14Lap\\
\cline{2-3}
 & Acc → ARS & Acc → ARS \\
\hline
BERT & 83.04 → 54.82 (↓28.22)  & 77.59 → 50.94 (↓26.65) \\
\hline
RGAT with Berkeley & 84.41 → 56.54 (↓27.87) & 78.09 → 51.37 (↓26.72) \\
RGAT with CoreNLP & 83.86 → 55.76 (↓28.10) & 78.12 → 52.27 (↓25.85) \\
RGAT with Stanza & 84.53 → 56.34 (↓28.19) & 77.99 → 51.20 (↓26.79) \\
\hline
\textbf{\methodname{}} & 85.16 → \textbf{57.46} (↓27.70) & 80.0 → \textbf{52.90} (↓27.10) \\
\bottomrule
\end{tabularx}
\end{threeparttable}
}
\caption{Comparison of \methodname{} model to the single dependency tree based models and BERT model in terms of Aspect Robustness
Score (ARS), on ARTS. 
}
\label{table:robustness}
\end{table}

\xhdr{Aspect Robustness.}
To study the aspect robustness of the \methodname{} model, we test our model on the Aspect Robustness Test Set (ARTS) datasets proposed by \citet{xing2020tasty} (see Table \ref{table:RobustnessData} for statistics). The datasets enrich the original 14Lap and 14Rest datasets following three strategies: reverse the sentiment of the aspect term; reverse the sentiment of the non-target terms with originally the same sentiment as target term; generate more non-target aspect terms that have opposite sentiment polarities to the target one.
They propose a novel metric, Aspect Robustness Score (ARS), that counts the correct classification of the source example and all its variations generated by the above three strategies as one unit
of correctness. 

We compare three single dependency tree models with the \methodname{} model in Table \ref{table:robustness}. We directly evaluate the models trained on the original SemEval datasets on ARTS without further tuning. The results indicate that the \methodname{} model shows better aspect robustness than single dependency tree and BERT models. 

\section{Conclusion}
We propose a simple yet effective graph-ensemble technique\peng{, \methodname{},} to combine multiple dependency trees \peng{for aspect-level sentiment analysis}. 
\peng{By taking the union of edges from different parsers, \methodname{} allows graph neural model to be robust to parse errors without additional parameters or computational cost.
With different edge types to capture the original syntactic dependency in parse trees, our model outperforms previous state-of-the-art models, single-parse models, as well as traditional ensemble models on three aspect-level sentiment classification benchmark datasets.}
\section*{Acknowledgement}
This work was supported by the National Key R\&D Program of China under Grant No.2020AAA108600.

\bibliography{anthology}
\bibliographystyle{acl_natbib}

\end{document}